\pdfoutput=1

\documentclass[11pt]{article}

\usepackage[preprint]{acl}

\usepackage{times}
\usepackage{latexsym}

\usepackage[T1]{fontenc}

\usepackage[utf8]{inputenc}

\usepackage{microtype}

\usepackage{inconsolata}

\usepackage{graphicx}
\usepackage{amsmath}
\usepackage{amssymb}
\usepackage{booktabs}
\usepackage{adjustbox}
\usepackage{tabularx}
\usepackage{multicol}
\usepackage{multirow}
\usepackage{subcaption}

\title{\textsc{ZigzagAttention}: Efficient Long-Context Inference with\\Exclusive Retrieval and Streaming Heads}

\author{Zhuorui Liu\textsuperscript{1} \quad
  Chen Zhang\textsuperscript{1} \quad
  Dawei Song\textsuperscript{1,}\thanks{Corresponding author.} \\
  \textsuperscript{1}Beijing Institute of Technology \\
  \texttt{\{zhuoruiliu1216,chenzhang9702\}@outlook.com,dwsong@bit.edu.cn} \\
  }

\begin{document}
\maketitle
\begin{abstract}
With the rapid development of large language models (LLMs), handling long context has become one of the vital abilities in LLMs. Such long-context ability is accompanied by difficulties in deployment, especially due to the increased consumption of KV cache. There is certain work aiming to optimize the memory footprint of KV cache, inspired by the observation that attention heads can be categorized into retrieval heads that are of great significance and streaming heads that are of less significance. Typically, identifying the streaming heads and and waiving the KV cache in the streaming heads would largely reduce the overhead without hurting the performance that much. However, since employing both retrieval and streaming heads in one layer decomposes one large round of attention computation into two small ones, it may unexpectedly bring extra latency on accessing and indexing tensors. Based on this intuition, we impose an important improvement to the identification process of retrieval and streaming heads, in which we design a criterion that enforces exclusively retrieval or streaming heads gathered in one unique layer. In this way, we further eliminate the extra latency and only incur negligible performance degradation. Our method named \textsc{ZigzagAttention} is competitive among considered baselines owing to reduced latency and comparable performance. 

\end{abstract}

\section{Introduction}
In recent years, large language models (LLMs)~\cite{dubey2024llama, liu2024deepseek} have demonstrated significant potential across diverse domains~\cite{vicuna2023}. 
However, the generation process of LLMs is inherently sequential. The sequential nature inevitably leads to substantial serving latency, particularly in scenarios involving long contexts. 

The primary challenge of serving LLMs for  long-context applications lies in the $O(n^2)$—where $n$ denotes the sequence length—complexity of attention~\cite{vaswani2017attention}. 
The inference can be divided into two phases, i.e., prefilling phase and decoding phase. Essentially, in the decoding phase, a linear increase in memory would be natural due to the use of the key-value (KV) cache technique, which stores intermediate representations of previously seen tokens to reduce latency. In long-context scenarios, the memory of the KV cache can even exceed that of the model itself~\cite{liu2023deja}. 

To address the memory burden imposed by KV cache, numerous approaches have been proposed to optimize the KV cache from various perspectives. 
Among these, DuoAttention~\cite{xiao2024duoattention} is a typical representative. DuoAttention intends to identify \textit{retrieval heads}~\cite{wu2024retrieval} that are of great importance for long-context modeling and \textit{streaming heads}~\cite{xiao2023efficient} that are of less importance, and predominantly waive the KV cache in the streaming heads.
In doing so, DuoAttention has preserved the long-context capabilities of LLMs while improving computational efficiency. 

However, DuoAttention requires processing attention computations twice separately for retrieval heads and streaming heads within one layer. 
Unfortunately, such separation necessitates additional memory accessing and introduces unwanted tensor indexing, leading to increased latency. This overhead becomes pronounced particularly along the expansion of context. Based on the intuition, we propose a valuable rearrangement of the retrieval and streaming heads. By enforcing either retrieval or streaming heads mutually exclusive across layers, we can perform one attention computation at each layer, thereby avoiding extra latency associated with redundant memory accessing and tensor indexing. 

On an extensive set of experiments ranging from LongBench to Needle-in-a-Haystack, our proposed method \textsc{ZigzagAttention} achieves competitive performance while significantly reduced latency.

\section{\textsc{ZigZagAttention}}

\subsection{Preliminary}

To identify retrieval and streaming heads in a LLM, DuoAttention firstly plugs an importance score $\alpha\in[0,1]$ onto each attention head, secondly employs a distillation-driven training on a synthetic dataset curated in the form of long-context passkey retrieval, and finally determines retrieval and streaming heads based on the descending order of the converged $\alpha$ values and a predefined quantile.

Specifically, $\alpha$ for each head is initialized to 1, and constrained to the range $[0,1]$. During the training, it performs attention computation twice in each forward pass: one using full attention (corresponding to retrieval head), and another using streaming attention (corresponding to streaming head). This is formalized as follows:
\begin{equation}
\label{eq:duoattn}
\begin{aligned}
    \textsf{attention}_{i,j} = \quad &\alpha_{i,j}\cdot\textsf{full\_attention}\quad +  \\
                             &(1-\alpha_{i,j}) \cdot \textsf{streaming\_attention}
\end{aligned}
\end{equation}
where $i$ and $j$ denote the layer index and the attention head index within a layer, respectively.
A synthetic dataset is used, with passkeys inserted at varying depths in the sequence, as the training task. The distillation-like training objective is formulated as follows:
\begin{equation}
    \mathcal{L}_{\sf dist} = \frac{1}{K} \sum_{k=1}^{K}\sum_{t=T-R+1}^{T}(\textbf{h}_{\sf full}^{(k)}[t] - \textbf{h}_{\sf mix}^{(k)}[t]) ^2
\end{equation}
where $K$ represents the dimension of hidden states, $T$ denotes the total sequence length, and $R$ means to the response length of the sequence. $\textbf{h}_{\sf full}$ and $\textbf{h}_{\sf mix}$ refer to the final hidden states from the standard full attention and the mixed attention computed in Equation~\ref{eq:duoattn}, respectively. To ensure sparsity in $\alpha$, an additional $L_1$ regularization term~\cite{tibshirani1996regression} is added:
\begin{equation}
    \mathcal{L}_{\sf reg} = \sum_{i=1}^{L}\sum_{j=1}^{H}|\alpha_{i,j}|
\end{equation}
where $L$ is the number of layers in the model, and $H$ is the number of attention heads per layer. The final loss function is formulated as:
\begin{equation}
    \mathcal{L}_{\sf duo} = \mathcal{L}_{\sf dist} + \lambda\mathcal{L}_{\sf reg}
\end{equation}
where $\lambda$ is a coefficient controlling the impact of the regularization term. After training, the attention heads are sorted based on their final $\alpha$ values. By specifying a custom sparsity quantile, the heads can be categorized as either full attention (retrieval heads) or streaming attention (streaming heads).

As we can observe in DuoAttention, if a quantile is defined to categorize attention heads, different attention heads are likely to coexist within one layer. This kind of allocation may introduce extra latency due to the need for separate computations for each type of attention head. 

\subsection{Transport Optimization}

To alleviate the need of two rounds of attention computations, we consider the most straightforward way to achieve so. That is, leveraging the converged $\alpha$ values from DuoAttention, and defining the transition from DuoAttention to \textsc{ZigzagAttention} a transport optimization problem. Provided that the original sparsity (or say the proportion of streaming heads) in DuoAttention is $s$, accordingly in \textsc{ZigzagAttention}, the number of layers corresponding to all streaming heads should be $p$ where $p/L=s$, and the number of layers corresponding to all retrieval heads should be $q=L-p$.  

In the transport optimization problem, there is a operation set $O={o_{i,j}}$ comprising of totally $L\cdot H$ operations need to be carried out, and three operations are defined: 1) maintaining the type of attention head $o^{(0)}$, 2) turning a retrieval head to streaming one $o^{(1)}$, and reversely turning a streaming head to retrieval one $o^{(2)}$. Ideally, shifting from a retrieval head to streaming head would lead to performance decline, while shifting from a streaming head would lead to performance boost. Thereby, the optimization objective is shown below:
\begin{equation}
\begin{aligned}
    \min_{o_{i,j}}\mathcal{L}_{\sf zigzag}&\quad s.t. \quad p+q=L \\
    \mathcal{L}_{\sf zigzag}&=\sum_{i=1}^{L}\sum_{j=1}^{H}\hat{\alpha}_{i,j} \\ \hat{\alpha}_{i,j}&=\begin{cases}
        0, \quad o_{i,j}=o^{(0)} \\
        \alpha_{i,j}, \quad o_{i,j}=o^{(1)} \\
        -\omega\cdot\alpha, \quad o_{i,j}=o^{(2)}
    \end{cases}
\end{aligned}
\end{equation}
Enumeratively, the number of possible combinations under the subjection $p+q=L$ is $\tbinom{f}{L}$. Once one of these combinations is used, then the operation set $O$ should also be determined. Since $\tbinom{f}{L}$ is computationally trackable, we empirically examine each of them one by one and uncover the one yielding the minimum. $\omega \in [0, 1]$ represents a scaling factor, which is determined through grid search to identify its optimal value.

\subsection{Fine-tuning for Enhanced Ability}
After optimization, we observe that while the performance is comparable to the baseline, optionally, fine-tuning with minimal training cost can further enhance performance on certain benchmarks, particularly retrieval tasks.

For fine-tuning, we adopt the previously used training scheme in DuoAttention and plug the layer-wise $\alpha_l$s onto layers rather than heads using the optimal combination from the aforementioned transport optimization. 
By leveraging these trained results, we can sparsify the model to achieve improved performance.

\section{Experiments}
\begin{figure}[ht]
    \centering
    \includegraphics[width=0.7\linewidth]{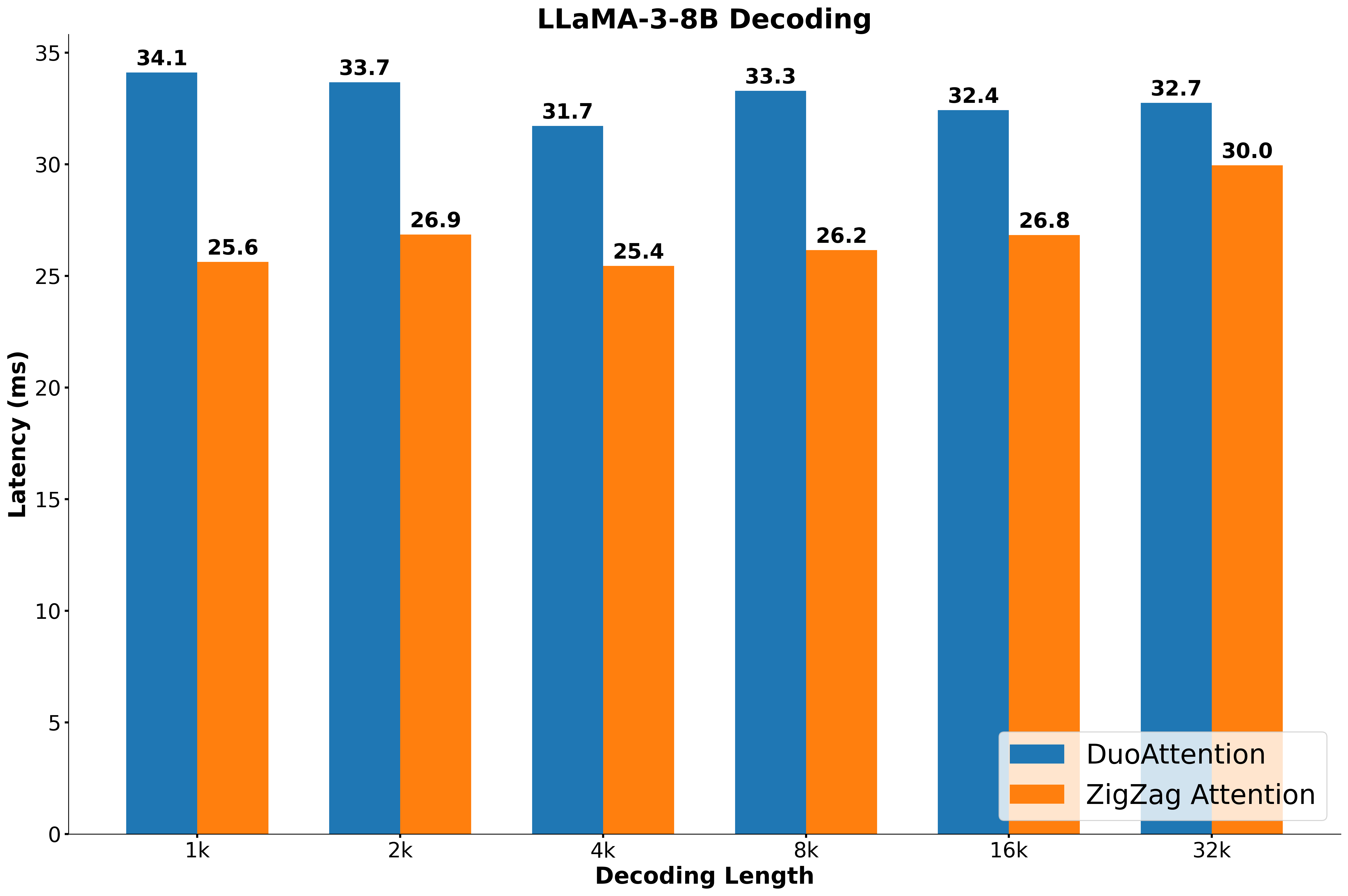}
    \caption{Per token decoding latency. The prefilling length here is set to 16k, and the decoding length varies from 1k to 32k.}
    \label{fig:efficiency_res}
\end{figure}
\begin{figure}[ht]
    \centering
    \includegraphics[width=0.7\linewidth]{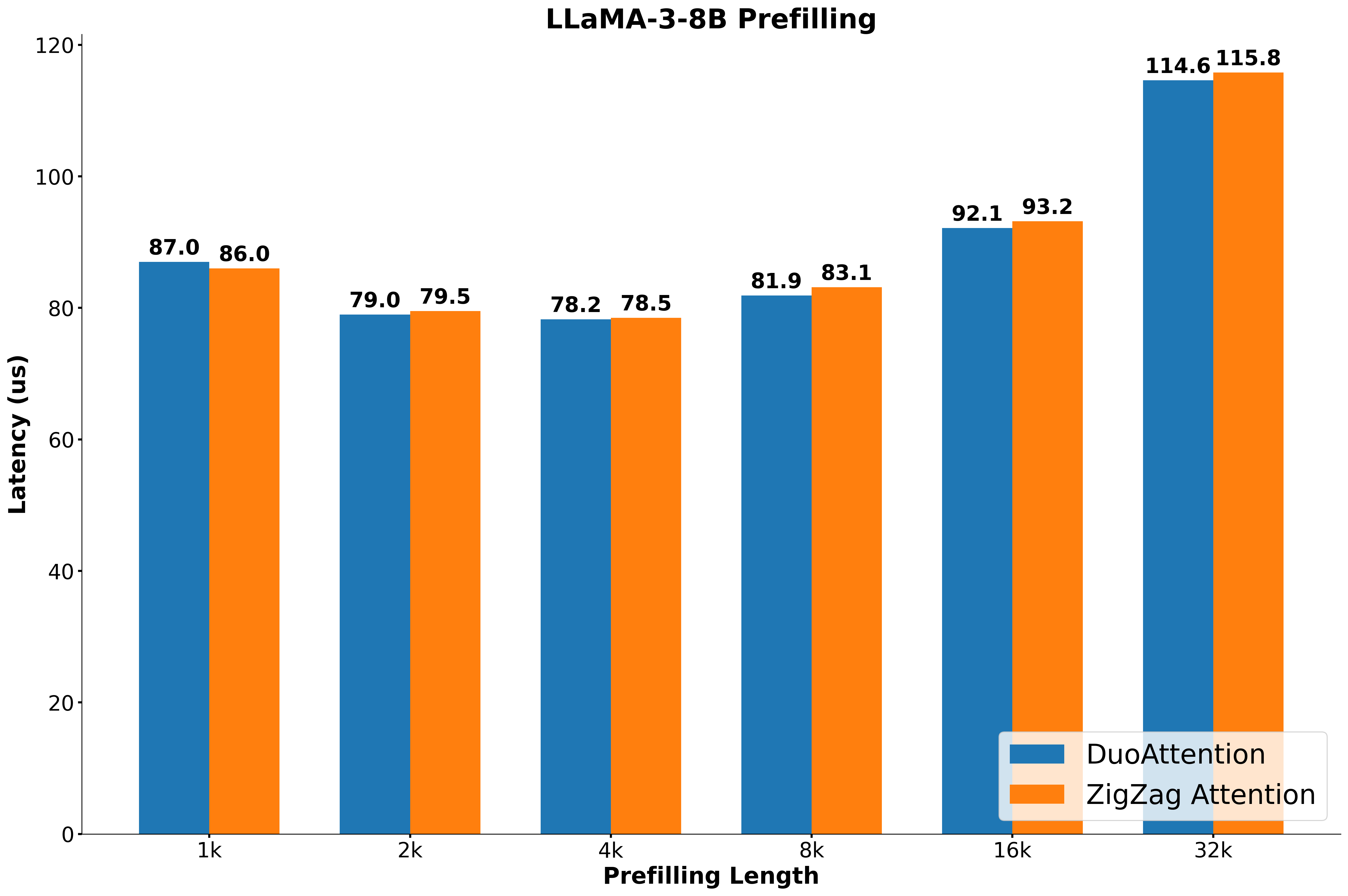}
    \caption{Per token prefilling latency. The decoding length here is set to 1k, and the prefilling length varies from 1k to 32k.}
    \label{fig:prefilling_res}
\end{figure}

\begin{table*}
    \centering
    \caption{Evaluation results on LongBench. Here "LM-3" refers to the results of LLaMA-3-8B long context extension version, "DA" refers to DuoAttention, "ZA" refers to ZigZag Attention.}
    \begin{adjustbox}{width=\textwidth,center}
    \begin{tabular}{l*{16}c}
        \toprule
        \multirow{2}{*}{Method} & \multicolumn{3}{c}{ Single-Document QA} & \multicolumn{3}{c}{ Multi-Document QA} & \multicolumn{3}{c}{ Summarization} & \multicolumn{3}{c}{ Few-shot Learning} & \multicolumn{2}{c}{ Synthetic} & \multicolumn{2}{c}{ Code} \\
        \cmidrule{2-17}
        & \rotatebox{50}{ NrtvQA} & \rotatebox{50}{ Qasper} & \rotatebox{50}{ MF-en} & \rotatebox{50}{ HotpotQA} & \rotatebox{50}{ 2WikiMQA} & \rotatebox{50}{ Musique} & \rotatebox{50}{ GovReport} & \rotatebox{50}{ QMSum} & \rotatebox{50}{ MultiNews} & \rotatebox{50}{ TREC} & \rotatebox{50}{ TriviaQA} & \rotatebox{50}{ SAMSum} & \rotatebox{50}{ PCount} & \rotatebox{50}{ PRe} & \rotatebox{50}{ Lcc} & \rotatebox{50}{ RB-p} \\
        \midrule
        LM-3 &  26.84 &  29.32 &  52.86 &  40.87 &  28.86 &  24.68 &  34.25 &  24.58 &  27.8 &  71.0 &  87.7 &  41.95 &  1.0 &  79.0 &  37.91 &  37.71 \\
        \midrule
        DA &  25.72 &  28.35 &  49.75 &  43.28 &  29.9 &  23.41 &  32.34 &  24.69 &  28.06 & 72.0 & 86.85 &  41.97 &  1.5 &  83.12 &  38.33 &  39.5 \\
        ZA &  22.53 &  23.7 &  49.89 &  38.53 &  23.61 &  21.21 &  30.62 & 24.16 &  27.12 &  71.0 &  82.22 &  40.85 & 1.0 & 85.0 &  45.38 &  44.7 \\
        \bottomrule

        \end{tabular}
        \end{adjustbox}
    \label{tab:longbench}
\end{table*}

\subsection{Settings}
We conduct experiments using the long-context extension version of the LLaMA-3-8B~\cite{dubey2024llama,gradientlongcontextllama3} model and evaluate its performance on both long-context and short-context benchmarks to ensure a comprehensive assessment. 
For long-context benchmarks, we select LongBench~\cite{bai2023longbench} and Needle-in-a-Haystack~\cite{needle}, while for short-context benchmarks, we choose MMLU~\cite{hendrycks2020measuring}, BBH~\cite{suzgun2022challenging}, and DROP~\cite{dua2019drop}. To evaluate efficiency, we test the model's performance across various combinations of prefilling and decoding lengths to minimize the impact of measurement errors. The primary training settings are aligned with those used in DuoAttention.

\subsection{Efficiency Results} 

The results are illustrated in Figure~\ref{fig:efficiency_res}, while DuoAttention demonstrates lower latency and better performance compared to the original model, \textsc{ZigZagAttention} achieves even lower latency across all decoding lengths. 
\textsc{ZigZagAttention} achieves up to 37\% acceleration in 1k context length. 
Figure~\ref{fig:prefilling_res} compares the per-token prefilling latency between \textsc{ZigZagAttention} and DuoAttention, since \textsc{ZigZagAttention} do not modify the prefilling stage, our method maintains normal prefilling speed. This confirms that \textsc{ZigZagAttention} does not introduce additional latency during the prefilling stage.
As for the time cost of the transport problem, the total time cost with our optimized method is around \textbf{7 minute}.

\subsection{Long Context Benchmark}
\begin{table}[ht]
    \centering
    \caption{The average scores on overall LongBench.}
    \begin{adjustbox}{width=0.25\textwidth,center}
    \begin{tabular}{lcc}
    \toprule
        Method & Budget & LongBench \\
        \midrule
        LM-3 & 100\% & 39.78 \\
        \midrule
        DA & 50\% & 39.45 \\
        ZA & 50\% & 38.44 \\
        \bottomrule
        
    \end{tabular}
    \end{adjustbox}
    \label{tab:longbench_overall}
\end{table}

For this evaluation, we applied a 50\% sparsity level for the LLaMA-3-8B model and set the sink size to 128 and window length to 256 for streaming attention. 

\paragraph{LongBench} 
The results for selected datasets are presented in Table~\ref{tab:longbench}, while average scores across all tasks are shown in Table~\ref{tab:longbench_overall}. 
From Table~\ref{tab:longbench}, importantly, there is no significant decline in metrics compared to the original model, demonstrating that \textsc{ZigZagAttention} can effectively manage long-context situations. 
As shown in Table~\ref{tab:longbench_overall}, \textsc{ZigZagAttention} scores are only marginally lower, compared to DuoAttention and the original model.

\begin{figure}[ht]
    \centering
    \includegraphics[width=0.5\linewidth]{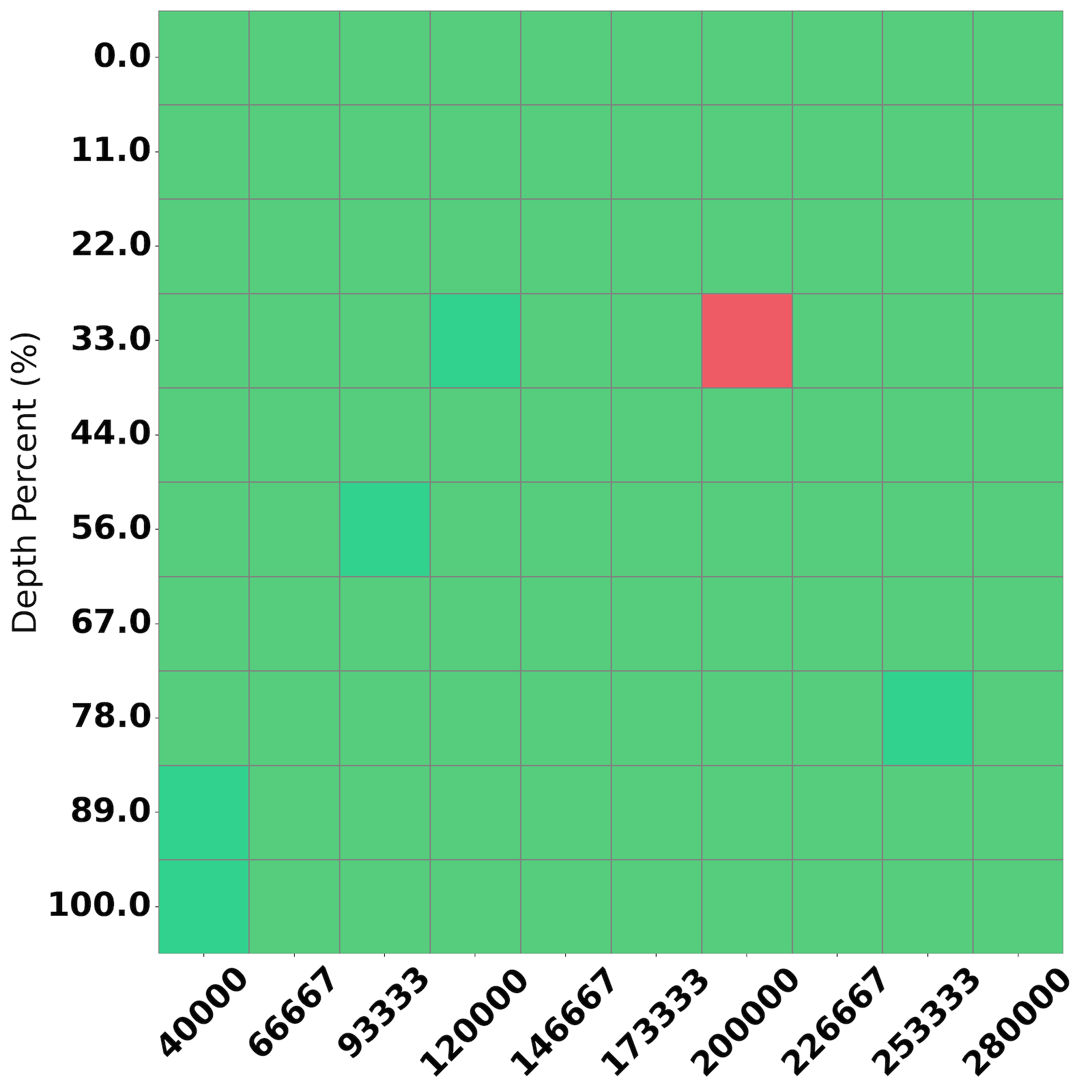}
    \caption{Results on NIAH varies from 40k to 280k.} 
    \label{fig:niah}

\end{figure}

\paragraph{Needle-in-a-Haystack (NIAH)} 
Figure~\ref{fig:niah} illustrates that \textsc{ZigZagAttention} performs exceptionally well across various context lengths ranging from 40k to 280k tokens. The results indicate that \textsc{ZigZagAttention} successfully discards unimportant KV caches during inference, without any performance degradation in complex long-context retrieval tasks. 

\subsection{Short Context Benchmark}
\begin{table}[ht]
    \centering
    \caption{Evaluation results on MMLU, BBH and DROP benchmarks.}
    \begin{adjustbox}{width=0.35\textwidth,center}
    \begin{tabular}{lcccc}
    \toprule
        \multirow{2}{*}{Method} & \multirow{2}{*}{Budget} & MMLU & BBH & DROP \\
        & & 5-shot & 3-shot & 3-shot\\
        \midrule
        LM-3 & 100\% & 62.31 & 41.95 & 44.18 \\
        \midrule
        DA & 50\% & 62.56 & 42.14 & 42.07 \\
        ZA & 50\% & 62.31 & 42.03 & 43.50 \\
        \bottomrule
        
    \end{tabular}
    \end{adjustbox}
    \label{tab:short_bench}
\end{table}
\begin{table}[ht]
    \centering
    \caption{The average scores on overall LongBench.}
    \begin{adjustbox}{width=0.25\textwidth,center}
    \begin{tabular}{lcc}
    \toprule
        $\omega$ & Budget & LongBench \\
        \midrule 
        $0.1$ & 50\% & 38.44 \\
        $0.5$ & 50\% & 37.59 \\
        $0.6$ & 50\% & 37.08 \\
        $0.7$ & 50\% & 37.05 \\
        $0.8$ & 50\% & 35.27 \\
        $0.9$ & 50\% & 36.02 \\
        \bottomrule
        
    \end{tabular}
    \end{adjustbox}
    \label{tab:ablation_longbench_overall}
\end{table}
As shown in Table~\ref{tab:short_bench}, \textsc{ZigZagAttention} demonstrates performance comparable to that of the base model LLaMA-3 across these important benchmarks. This indicates that the \textsc{ZigZagAttention} mechanism does not impair the model's basic capabilities. 

\subsection{Ablation Study}

\paragraph{Impact of $\omega$}

Changes in $\omega$ can alter the combination of layers, potentially affecting the model's performance in long-context situations and benchmarks.
Specifically, when $\omega$ is set to 0.2, 0.3, or 0.4, the final combinations are identical to those obtained with $w=0.1$. 
In Table~\ref{tab:ablation_longbench_overall}, it is evident that $w=0.1$ yields the optimal combination with the best performance across multiple tasks in long-context benchmarks.



\begin{figure}[ht]
    \centering
    \includegraphics[width=0.5\linewidth]{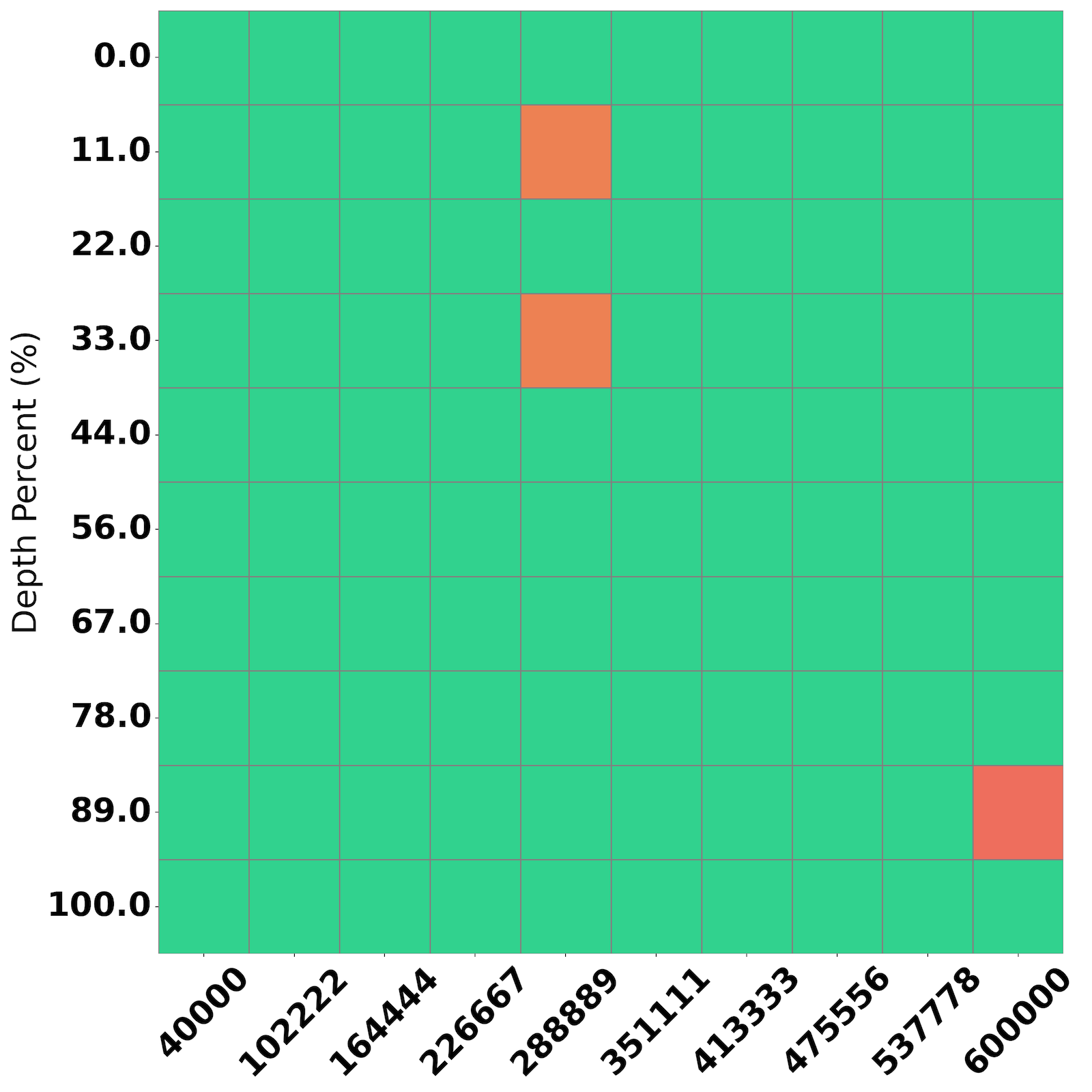}
    \caption{Results for after training in ZA. We successfully extent the context length to 600k.}
    \label{fig:ablation_niah}
\end{figure}

\paragraph{Context Length Extension with Fine-tuning}


As shown in Figure~\ref{fig:ablation_niah}, the fine-tuning allowed us to extend the context length from 280k tokens to 600k tokens with minimal additional training. 

\section{Conclusion}

In this paper, we introduced \textsc{ZigZagAttention}, a method built upon DuoAttention and designed to address the challenges of handling long-context situations. 
Our results demonstrate that \textsc{ZigZagAttention} achieves performance comparable to the original model, indicating that it can significantly lower latency without degrading model capabilities.


\section*{Limitations}
In this paper, we propose \textsc{ZigzagAttention} to accelerate model inference. However, the current method has certain limitations. In terms of efficiency, the speedup ratio decreases for longer decoding lengths compared to shorter ones, resulting in less significant performance improvements. For retrieval tasks, \textsc{ZigzagAttention} achieves an overall high score but still exhibits performance degradation relative to other methods. Nevertheless, these limitations highlight key areas for further analysis and provide a clear direction for future research.
\bibliography{custom}

\clearpage
\appendix



\end{document}